\documentclass[conference]{IEEEtran}

\usepackage{times}
\usepackage{latexsym}
\usepackage[T1]{fontenc}
\usepackage[utf8]{inputenc}
\usepackage{microtype}
\usepackage{booktabs}
\usepackage{tabularray}
\usepackage{amsmath}
\usepackage{graphicx}
\usepackage{hyperref} 
\usepackage{algorithm}
\usepackage{algorithmic}
\usepackage{multirow}
\usepackage{tikz}
\usepackage{makecell}
\usepackage{tikz}
\usepackage{subcaption}
\def\checkmark{\tikz\fill[scale=0.4](0,.35) -- (.25,0) -- (1,.7) -- (.25,.15) -- cycle;}

\ifCLASSOPTIONcompsoc
  \usepackage[nocompress]{cite}
\else
  \usepackage{cite}
\fi
\ifCLASSINFOpdf
\else
\fi
\makeatletter 
\newcommand{\linebreakand}{%
  \end{@IEEEauthorhalign}
  \hfill\mbox{}\par
  \mbox{}\hfill\begin{@IEEEauthorhalign}
}
\makeatother 

\begin{document}
%

\title{Robust Stance Detection: Understanding\\Public Perceptions in Social Media} 

\author{\IEEEauthorblockN{Nayoung Kim\IEEEauthorrefmark{1},
David Mosallanezhad\IEEEauthorrefmark{2}, Lu Cheng\IEEEauthorrefmark{3},
Michelle V. Mancenido\IEEEauthorrefmark{1} and Huan Liu\IEEEauthorrefmark{1}}
\IEEEauthorblockA{\IEEEauthorrefmark{1}Arizona State University, \IEEEauthorrefmark{2}NVIDIA, \IEEEauthorrefmark{3}University of Illinois Chicago
\\
Email: \IEEEauthorrefmark{1}\{nkim48, mmanceni, huanliu\}@asu.edu,
\IEEEauthorrefmark{2}dmosallanezh@nvidia.com,
\IEEEauthorrefmark{3}lucheng@uic.edu}}

\newcommand{\model}{\textsc{STANCE-C3}}
\newcommand{\David}{\textcolor{black}}
\newcommand{\mickey}[1]{\textcolor{blue}{#1}}

\maketitle 

\begin{abstract}
The abundance of social media data has presented opportunities for accurately determining public and group-specific stances around policy proposals or controversial topics. In contrast with sentiment analysis which focuses on identifying prevailing emotions, stance detection identifies precise positions (i.e., supportive, opposing, neutral) relative to a well-defined topic, such as perceptions toward specific global health interventions during the COVID-19 pandemic. Traditional stance detection models, while effective within their specific domain (e.g., attitudes towards masking protocols during COVID-19), often lag in performance when applied to new domains \emph{and} topics due to changes in data distribution. This limitation is compounded by the scarcity of domain-specific, labeled datasets, which are expensive and labor-intensive to create. A solution we present in this paper combines counterfactual data augmentation with contrastive learning to enhance the robustness of stance detection across domains and topics of interest. We evaluate the performance of current state-of-the-art stance detection models, including a prompt-optimized large language model, relative to our proposed framework succinctly called \model~(domain-adaptive Cross-target STANCE detection via Contrastive learning and Counterfactual generation). Empirical evaluations demonstrate \model's consistent improvements over the baseline models with respect to accuracy across domains and varying focal topics. Despite the increasing prevalence of general-purpose models such as generative AI, specialized models such as \model~provide utility in safety-critical domains wherein precision is highly valued, especially when a nuanced understanding of the concerns of different population segments could result in crafting more impactful public policies. 

\end{abstract}


%
\IEEEpeerreviewmaketitle

\section{Introduction}
\label{s:introduction}


The influence of public perceptions on policy development and societal outcomes is well-established in the context of public health~\cite{zhang2017we}, legislation~\cite{nababan2022twitter}, environmental sustainability~\cite{dash2023sustainable}, and other community-focused sectors. While policymakers from previous generations relied on expensive surveys or town halls to gauge public opinion toward a specific issue, the meteoric rise of social media has facilitated the swift and cost-effective collection of user-generated insights, which are voluntarily and freely provided by netizens. At the height of COVID-19, for example, social media data and natural language processing (NLP) models were leveraged to establish associations between public support for non-pharmaceutical interventions and the decreased spread of the virus \cite{agusto23}. 

NLP models that infer the collective attitude or stance toward specific targets using corpora of user-provided data from online forums (e.g., X) are called \emph{stance detection} models. Unlike sentiment analysis, which detects general emotional responses toward a topic, stance detection models identify positions toward a target by analyzing the context, linguistic nuances, and even the implied meanings within the text. For instance, consider an online post stating, ``\emph{I hate having to wear a mask every day, but I totally agree it's essential to stop the spread of COVID-19}." Sentiment analysis might classify this as negative due to the expressions of dislike and discomfort. However, despite the negative emotional tone, the stance towards the target (mask-wearing as a COVID-19 health measure) is clearly supportive. Consequently, stance detection offers significant opportunities to leverage social media data for accurately identifying public stances towards various initiatives. This can substantially enhance policy-making, refine strategies that impact the public, and contribute to a more informed public dialogue~\cite{agusto2023impact}.

\begin{figure}[!t]
    \centering
    \begin{subfigure}[b]{0.16\textwidth}
        \centering
        \includegraphics[width=\textwidth]{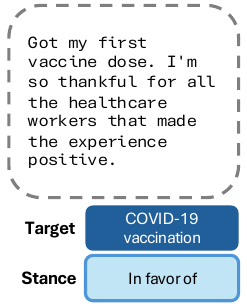}
        \caption{Source Domain}
    \end{subfigure}
    \hspace{1em}
    \begin{subfigure}[b]{0.16\textwidth}
        \centering
        \includegraphics[width=\textwidth]{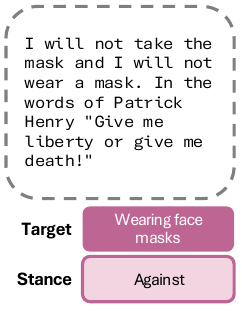}
        \caption{Target Domain}
    \end{subfigure}
    
    \caption{{\footnotesize Examples of domain-adaptive cross-target stance detection. (a) Tweets from source domain are collected using target-related keywords/hashtags (e.g., \textit{get vaccinated}) from January 1st, 2020 to August 23rd, 2021, whereas (b) tweets from target domain are collected with target-related hashtags (e.g., \textit{\#MasksSaveLives}) from February 27th, 2020 to August 20th, 2020.}}
    \label{example} 
\vspace{-10pt}
\end{figure}
  

\begin{figure*}[ht!]\vspace{-15pt}
    \centering
    \includegraphics[scale=0.78]{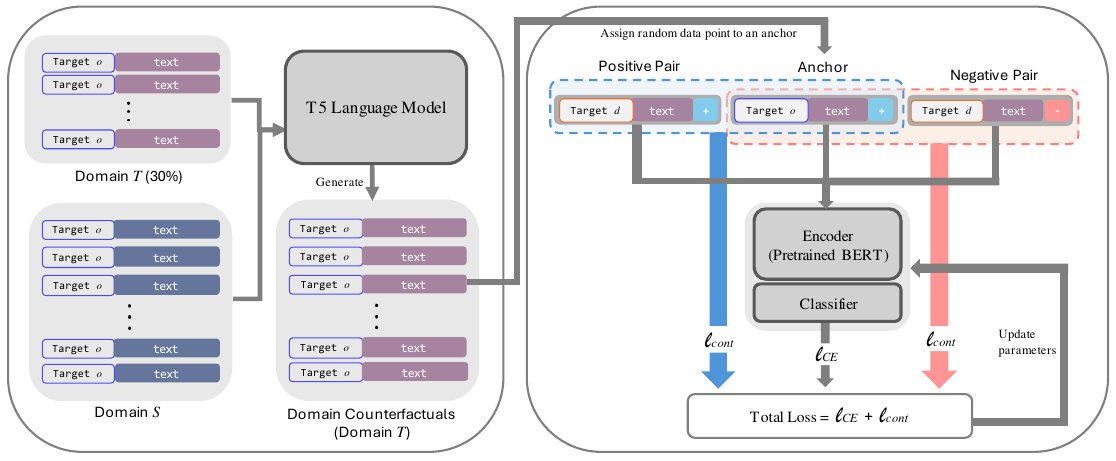}
    \caption{{\footnotesize The STANCE-C3 architecture consists of two key components: a counterfactual data generation network and a contrastive learning network. The counterfactual network (left) is built on a T5-based model and generates training examples that maintain sentence structure but has increased diversity in semantic context. The contrastive learning network (right) then uses the augmented dataset to learn cross-target representations by minimizing distances between examples with the same stance (positive pairs) and maximizing distances between examples with different stance (negative pairs). This approach helps acquire domain-invariant features, improving stance detection across targets.} }
    \label{fig:proposed}

\end{figure*}

One of the significant challenges in leveraging social media data for stance detection is the limited adaptability of existing models across different domains. This limitation is particularly acute in rapidly changing situations, where shifting public opinion across various domains (e.g., different time points) can significantly influence responses to crises, such as public health emergencies. The social cost of misinformation and the urgent need for responsive public opinion analysis across diverse domains highlight a pressing challenge: developing stance detection models that are both robust and adaptable to different contexts.

Current state-of-the-art (SOTA) stance detection models are largely focused on improving the model's performance within a single domain. Yet, these models often underperform when applied to domains distinct from their training datasets~\cite{kaushal2021twt}. We attribute this performance degradation to several factors: (1) Attention bias toward events related to the given target, a phenomenon we will illustrate with experiments in later sections; (2) The absence of target words in stance prediction during the training phase; and (3) The model's reliance on auxiliary words (i.e., words that are not necessarily the target or stance) specific to a given domain. Further, the limited availability of training data restricts these models' effectiveness across multiple domains and targets. Although Large Language Models (LLMs) have gained popularity, their resource-intensive and general-purpose nature pose issues in their usefulness for stance detection. LLMs demand substantial computational resources for training and inference, making them financially and environmentally costly. For specific tasks, smaller models can achieve comparable or even better performance without the resource burden~\cite{Turc2019WellReadSL,hoffmann2022empirical}. Additionally, human factors significantly influence the inference process when using LLMs; for example, it is well-established that different prompting techniques could yield widely varying results~\cite{zamfirescu2023johnny}. 

Stance detection models can also be differentiated based on the number of intended targets. Single-target models identify positions toward a specific, fixed target~\cite{mohammad2016semeval}, whereas cross-target models extend the model's versatility to multiple targets~\cite{augenstein-etal-2016-stance}. Approaches for single-target stance detection use training sets that only include labeled data related to the specific target. In contrast, cross-target methods address the challenge of the sparseness in target-specific labeled data, leveraging additional information about the new target for accurate stance predictions.

With social media serving as the main platform where people express their opinions and grievances, the development of robust stance detection methods is crucial for accurately gauging public opinion, deterring misinformation, and facilitating constructive online dialogue. Single-target models often struggle to adapt to the quickly evolving landscape of public discourse, where topics such as climate change, social justice, and public health are deeply interwoven in their influence and digital presence. Cross-target stance detection models offer a more comprehensive understanding of public opinions across related topics, eliminating the need to rebuild models for each new issue. This not only makes AI tools more responsive and adaptable but also improves their ability to provide real-time insights across a wide range of issues. Such capabilities are particularly valuable in fields like policy making and public health, where timely and informed decision-making is crucial. For example, a cross-target model could analyze opinion on related non-pharmaceutical health interventions like vaccination and mask-wearing, despite varying domains and data sources, enabling more effective responses to public health crises. \autoref{example} shows an example of a domain-adaptive cross-target stance detection problem, illustrating differences in domains by search query and data acquisition time periods, while targets differ in topical content (e.g., \emph{vaccination} vs. \emph{mask-wearing}). Other attributes of the sampling process could account for distinctions among domains and targets, such as data sources (e.g., social media vs. news), content genre (e.g., political vs. gossip news), or publication type (e.g., opinion pieces vs. reportage).

In this paper, we introduce \model~as a domain- and target-robust model that uses Counterfactual Data Augmentation (CDA)~\cite{lu2020gender}, an approach that has been shown to decrease bias and improve the robustness of neural network-based models~\cite{wang2021robustness}. We combine CDA with text style transfer methods~\cite{10.1162/coli_a_00426} to facilitate effective text transfer from one domain to another. These language model-based data augmentation techniques are expected to enrich existing datasets, improving the performance of stance detection models across multiple domains.

Previous work on stance detection addressed domain robustness using adversarial training and incorporating the target word as an input to the model. However, this approach often led to overfitting on the specific target, restricting its robustness in performance for other targets. In \model, we modify a self-supervised technique called contrastive learning~\cite{chen2020simple} to extract both shared and distinct high-level features to capture specific characteristics of statement-target pairs. We incorporate components such as cosine similarity and negative pair loss to direct the model to minimize the distance between pairs with the same stance labels and maximize the distance between pairs with different stance labels. By requiring fewer manually annotated data points and combining CDA during the pre-processing stage, \model~ensures that the training set is optimally constructed for a cross-domain and cross-target setting (i.e., a sufficient balance between data points that exhibit similarities and differences, allowing the model to infer salient features that are useful for stance detection) while avoiding overfitting.

The major contributions of this work are threefold: firstly, it addresses the problem of domain-adaptive cross-target stance detection, offering a novel approach to train models effectively even when data on the target domain and subject are limited. Secondly, it tests the potential of a modified contrastive loss learning method as a more effective alternative for improving model robustness in classification tasks. Lastly, it demonstrates \model's efficacy and improvement over existing SOTA stance detection models through rigorous experiments and ablation studies on real-world datasets. 

\section{Related Work}
\label{s:related_work}

\model~is proposed in this paper to improve existing models for stance detection and robustness across domains. In this section, we review current NLP SOTA models for stance detection and domain adaptation. 

\subsection{Stance Detection}
There are two general categories of distinctions among stance detection models. Firstly, models are distinguished on how stance features are modeled. \textit{Content-level} approaches use linguistic features such as topics \cite{GOMEZSUTA2023119046} or targets \cite{wei2019modeling} alongside sentiment information \cite{sobhani-etal-2016-detecting}. \textit{User-level} approaches focus on user-related attributes interactions \cite{Rashed_Kutlu_Darwish_Elsayed_Bayrak_2021}, information \cite{benton-dredze-2018-using}, and timelines \cite{ZHU2020102031}. Hybrid models combine content- and user-level features for post representations \cite{10.3233/JIFS-179895}. In this paper, we focus on content-level modeling with target embedding using BERT due to its practicality in data acquisition. 

Secondly, stance detection models differ concerning the specificity of the target of interest. While many previous works focus on specific single-target scenarios \cite{Darwish_Stefanov_Aupetit_Nakov_2020}, recent studies have considered cross-target \cite{wei2019modeling}, multi-target \cite{li2021multi}, and few-shot or zero-shot stance detection \cite{liu2022target}. While both cross-target and zero-shot stance detection models are trained on one or more source targets and used on previously unseen targets \cite{liang2021target}, zero-shot methods can predict stances for unseen targets that are not necessarily related to the target in the training set~\cite{allaway-mckeown-2020-zero}. Both attempt to extract transferable knowledge from sources using methods such as hierarchical contrastive learning~\cite{liang2022zero}. \model~uses a similar but modified approach to the contrastive loss during training.

\subsection{Domain Adaptation}
As a category of transfer learning, domain Adaptation (DA) leverages agnostic or common information between domains with the target task remaining constant \cite{daume2006domain}. There are several DA setups, including unsupervised domain adaptation (UDA) where both labeled source data and unlabeled target data are available \cite{liu2022deep}, semi-supervised domain adaptation (SSDA) where labeled source data and a small number of labeled target data are accessible \cite{li2020online}, and supervised domain adaptation (SDA) where both labeled source and target data are available during training \cite{lin-lu-2018-neural}. Our focus is on SSDA, where a smaller set of labeled target data is used in the training process. 

Domain adaptive models are categorized by their focus on the model, training data, or both. Model-centric approaches aim to construct domain-agnostic model structures, such as in Deng et al.~\cite{deng2022domain}. Data-centric methods enhance robustness by leveraging on data attributes using techniques such as the incorporation of loss functions based on inter-domain distances \cite{guo2020multi} or the employment of pre-training \cite{gururangan-etal-2020-dont}. Finally, hybrid approaches combine model- and data-centric approaches \cite{hardalov-etal-2021-cross}. Many of these DA methods grapple with spurious correlations from differing training and test dataset distributions, thus limiting their effectiveness. Our hybrid approach addresses this by using domain counterfactual examples, which minimizes the impact of domain-invariant features in cross-domain classification.

\section{Problem Statement}
\label{s:problem_satement}

Suppose we have two sets of instances $\mathcal{D}_\mathcal{S}$ and $\mathcal{D}_\mathcal{T}$ from source domain $\mathcal{S}$ and target domain $\mathcal{T}$, respectively. The set $\mathcal{D}_\mathcal{S} = {(r_\mathcal{S}, t^o, y_\mathcal{S})}^N$ contains \textit{N} annotated instances from a source domain $\mathcal{S}$ where each instance $r_\mathcal{S}$ is a text (e.g., tweet) consisting of a sequence of \textit{k} words, and $y_\mathcal{S}$ is the stance toward a target $t^o$ associated with each annotated instance. Another set $\mathcal{D}_\mathcal{T} = (r_\mathcal{T}, t^d)^{N'}$ is a set of \textit{N'} instances from target domain $\mathcal{T}$ with the destination target $t^d$. The goal of the domain-adaptive cross-target stance detection is to learn the stance classifier $F$ that uses the features of the source domain $\mathcal{S}$ and target words $t^o$ in $\mathcal{D}_\mathcal{S}$ and predict the stance label $y_\mathcal{T}$ of each text $r_\mathcal{T}$ in $\mathcal{D}_\mathcal{T}$. We formally define the problem as:

\noindent\textbf{Definition} Given statements from two separate datasets $\mathcal{D}_\mathcal{S}$ and $\mathcal{D}_\mathcal{T}$, and corresponding targets $t^o$ and $t^d$ from the source $\mathcal{S}$ and target $\mathcal{T}$ domains, respectively, learn a domain and target-agnostic text representation using $\mathcal{D}_\mathcal{S}$ and a small portion of labeled $\mathcal{D}_\mathcal{T}$ that can be classified correctly by the stance classifier $F$.

\section{Proposed Model}
\label{s:proposed_model}

In this section, we describe our proposed model \model. As illustrated in \autoref{fig:proposed}, the architecture consists of two main components: (1) a domain counterfactual generator, which is trained with input representations of the labeled samples from the source domain and a small portion from the target domain, to generate a set of domain counterfactuals. These generated counterfactuals enrich the knowledge of the target domain during the stance classification training phase. (2) These samples are then combined with the original dataset (i.e., source domain dataset with a small portion of the target domain dataset) to train a BERT-based stance classifier. This training utilizes contrastive learning to develop a domain-adaptive, cross-target stance detection classifier. We partition a small subset (i.e., 50) of randomly selected samples from the combined datasets into positive and negative pairs based on their stance, regardless of the target. This contrastive loss approach enhances the method's ability to generalize stance detection to unseen targets. The following subsections provide a detailed explanation of these two stages.

\subsection{Domain-adaptive Single-target}
One major challenge in domain-adaptive stance detection is that collecting samples for a target domain can be expensive and usually requires human annotations. Even with annotated data, modeling a domain-adaptive stance detection is difficult due to the complexity of the dataset which contains far more distinct features over domains~\cite{schiller2021stance}. These features often form spurious correlations within a specific domain and make the models brittle to domain shift. For instance, the presence of question marks in a specific stance or highly negative sentiment associated with a particular stance has been demonstrated to lead to a performance decrease~\cite{kaushal2021twt}.

To address the aforementioned challenges and control these variants on the domain-adaptive stance detection problem, we adapt counterfactual concepts to the source domain. A domain-counterfactual example is defined as a result text of intervening on the domain variable of the original example and converting it to the target domain while keeping other features (e.g., overall structure of the sentence) fixed. Given an example data from the COVID-19 vaccination domain (source domain), the generator first recognizes the terms related to the domain. Then, it intervenes on these terms, replacing them with text that links the example to the wearing face mask domain (target domain) while keeping the stance. We utilize source domain instances in combination with a small portion of target domain instances to incorporate target domain information in generating the counterfactuals specific to the target domain. These counterfactuals are then combined as input for the subsequent stage. Following the work of ~\cite{calderon2022docogen}, we adopt a T5-based language model to generate counterfactuals for a given target domain $\mathcal{T}$. In this approach, we train the T5-based LM to convert samples from source domain $\mathcal{S}$ to a given target domain $\mathcal{T}$. 

Generating domain counterfactuals consists of two main steps - domain corruption and reconstruction. For domain corruption, we get the masking score of all n-grams $w$  ($n \in \{1, 2, 3\}$) in a dataset $\mathcal{D}_\mathcal{S}$ following~\cite{calderon2022docogen}. The affinity of $w$ to domain $\mathcal{S}$ is defined as $\rho(w,\mathcal{S}) = P(\mathcal{S}|w)(1-\frac{H(\mathcal{S}|w)}{\log N})$, where $H(\mathcal{S}|w)$ is the entropy of $\mathcal{S}$ given each n-gram $w$ and $N$ is the number of unlabeled domains we know. The final masking score given source domain $\mathcal{S}$ and target domain $\mathcal{T}$ is $mask(w, \mathcal{S}, \mathcal{T}) = \rho(w,\mathcal{S}) - \rho(w,\mathcal{T})$. A higher score implies the n-gram is highly related to the source domain and distant from the target domain. 

The next step involves reconstructing the masked examples $M(x)$ by concatenating them with a trainable embedding matrix, or domain orientation vectors. These vectors are embedded with domain-specific information including the nuances, vocabulary, and style of a particular domain. They are initialized with the embedding vector of each domain name and representative words from that domain. The concatenated matrix is then trained using a T5 architecture. Note that $\mathcal{T} = \mathcal{S}$ and $mask(w, \mathcal{S}, \mathcal{T}) = 0$ during the training process. Given the target domain and its orientation vectors in the test phase, the trained model generates domain counterfactual $x'$.


\subsection{Domain-adaptive Cross-target}
One of the main challenges in detecting stances across different targets is the variation in data distribution for each target, even within the same domain. For instance, the word \emph{WHO} may appear more frequently in the domain where the target is \emph{COVID-19 vaccination} than in another domain where the target is \emph{Donald Trump}~\cite{wang2020unseen}. The frequent co-occurrences of specific target words with certain instances make the model biases the model's learning process. Therefore it is necessary to identify an effective way to model transferable knowledge. To learn target-invariant features in instances from both the origin $t^o$ and destination $t^d$ targets, we use a simple yet effective supervised contrastive learning ~\cite{khosla2020supervised} approach. This loss function aims to enhance the separability of different classes by maximizing the distance between an anchor and negative samples.

In our problem setting, considering samples with different stance targets $t^o$ and $t^d$, the goal is to reduce the distance between samples sharing the same stance label and increase it for those with differing labels. Note that contrastive learning is applied to all pairs, regardless of whether the targets in the pair are equivalent or not. This approach creates a representation for the stance classifier that indicates the relation between the statement and the stance target $t$. We introduce a modified version of the supervised contrastive learning tailored to better address the nuances of our problem setting.

\small
\begin{align} \label{contloss}
& \ell_{\mathrm{cont}} = \sum_{i\in I} \frac{-1}{|P(i)|} \sum_{p\in P(i)} \log \frac{\exp(z_i \cdot z_p / \tau) \cdot sim(z_i,z_p)}{\sum_{a\in A(i)} \exp(z_i \cdot z_a/\tau)}  \notag \\ \nonumber
& \begin{aligned}[t] + \sum_{i\in I} \frac{1}{|N(i)|} \sum_{n\in N(i)} \log\frac{\exp(z_i \cdot z_n / \tau) \cdot sim(z_i,z_n)}{\sum_{a\in A(i)} \exp(z_i \cdot z_a/\tau)} \nonumber \end{aligned} 
\end{align}
\normalsize

The loss takes a batch of samples $i\in I$ as an ``anchor'' from the training set where each sample is mapped to a vector $z_i$. Each anchor allows for multiple positive vectors $z_p$ that have the same label as an anchor ($y_j=y_i$) in addition to multiple negative vectors $z_n$ ($z_n\in A(i)$, $y_n \neq y_i$). $P(i)$ and $N(i)$ represent the indices of all positive and negative vectors, respectively. $\tau$ shows the temperature parameter that affects the distance of instance discrimination. The notation $sim(\cdot)$ refers to the cosine similarity. Here, we introduce the second term to balance the weight of positive and negative pairs. According to our observations on experiments involving different variants of the modified contrastive loss, we discovered that weighing the original loss with text similarity improves the model's generalization ability across different targets. Note that anchor \textbf{$z_i$} is from target $t^o$ but \textbf{$z_p$} and \textbf{$z_n$} are from both $t^d$ and $t^o$ for target-agnostic feature learning. 

\noindent\textbf{Optimization Algorithm}
The training process of \model~for domain-adaptive cross-target scenario has two stages\footnote{Source code and data will become publicly available upon acceptance. Partial source code and data are available at \href{https://osf.io/26rfc/?view\_only=b6e3bfac71444eb0be0aaad0f671a720}{https://osf.io/26rfc/?view\_only=b6e3bfac71444eb0be0aaad0f671a720}}. In the first stage, the parameters of the T5-based LM are trained for conversion between source $\mathcal{S}$ and the target $\mathcal{T}$ domain following the work of~\cite{calderon2022docogen}. This stage enriches the input dataset to include more samples similar to the target domain. The next stage uses the contrastive learning approach to create a cross-target sentence representation. We randomly selected 50 anchors from $\mathcal{D}_\mathcal{S}$ and its domain counterfactuals, along with their positive and negative pairs from across all dataset except for generated counterfactuals. For each instance $x_i$, the overall loss is formulated as $\ell_{\mathrm{total}} = \lambda \ell_{\mathrm{cont}} + (1 - \lambda) \ell_{\mathrm{CE}}$ where $\ell_{\mathrm{CE}}$ stands for the cross-entropy loss for classification and ($\ell_{\mathrm{cont}}$) represents the contrastive loss. $\lambda$ determines the balance between the two loss components. Note that while the classification loss is computed over the entire training dataset, the contrastive loss is derived solely from the selected pairs mentioned previously.
The cross-entropy loss is calculated as $CE = -\sum_{i=1}^{k} y_i \log(p_i)$ for $k$ classes based on the true stance label $y$ and predicted stance probability $p$ for the $i^{th}$ class. Similar to \cite{khosla2020supervised} we set the temperature parameter $\tau=0.08$.


\section{Experiments}
\label{s:experiments}

\begin{table}[!t]
\scriptsize
\begin{center}
    \caption{Datasets' statistics: For COVID-19-Stance dataset, a significant number of tweets could not be downloaded due to Twitter restrictions and deletion of the tweets. In this context, the number of samples mentioned on the left refers to the actual sample size whereas the original data count is on the right.}
    \label{tab:table1}
    \begin{tabular}
    {p{0.14\linewidth}|p{0.3\linewidth}|p{0.1\linewidth}|p{0.25\linewidth}}
      \toprule 
      \textbf{Dataset} & \textbf{Target} & \textbf{Platform} & \textbf{Labels} \\ 
      \midrule 
      CoVaxNet  & COVID-19 Vaccination & Twitter & Pro (1,495,991), Anti (335,229) \\
      \midrule 
      COVID-19-Stance  & Wearing a face mask, Anthony S. Fauci, M.D., Keeping schools closed, Stay at home orders & Twitter & In-favor (2,421/43,799), Against (1,577/29,693) \\

      \bottomrule 
    \end{tabular}
  \end{center}
\vspace{-10pt}
\end{table}

The experiments conducted in this study investigate how changing the domains and targets affects the performance of several stance detection models, including \model. We also examine the effects of data augmentation and contrastive learning on performance accuracy in cross-domain and cross-target scenarios. More specifically, we aim to answer the following research questions: \textbf{Q1.} Does \model~ achieve better performance in comparison to other SOTA when trained on source domain $\mathcal{S}$ with some data from target domain $\mathcal{T}$, and tested on target domain $\mathcal{T}$ for target words in the source domain $\mathcal{S}$? \textbf{Q2.} Does \model~ achieve better performance in comparison to other SOTA when trained on a source domain $\mathcal{S}$ and tested on a target domain $\mathcal{T}$ for target words not in the source domain? \textbf{Q3.} What is the effect of changes in the model's parameters and components on its performance in the stance detection task? We consider the following scenarios in the experimental design: 
(1) single domain + single target, (2) cross-domain + single target, and (3) cross-domain + cross-target, where the targets and domains of the datasets have clear distributional differences. We address \textbf{Q1} by training and testing the baseline and proposed models under scenarios (1) and (2) where targets are the same for training and testing sets. For \textbf{Q2}, the models are evaluated under scenario (3) to compare and contrast the models' domain-adaptive and cross-target performance. Finally, we address \textbf{Q3} by performing ablation studies and parameter analysis. 
Notably, domain counterfactual generation is applied in all scenarios, and contrastive learning is employed only for scenario (3).

\begin{table}[!h] 
  \begin{center}
    \caption{Baselines targeted goals indicate the difference between the proposed approach \textsc{\model} and the baselines. }
    \label{tab:baselines}
    \scriptsize
    \resizebox{.8\linewidth}{!}{%
    \begin{tabular}{p{0.5\linewidth}|c|c}
      \toprule 
      \textbf{Model} & \textbf{Cross-Domain} & \textbf{Cross-Target}  \\ 
      \midrule 
      MoLE~\cite{hardalov-etal-2021-cross}  & \checkmark & \checkmark \\
      MTL~\cite{schiller2021stance}   & \checkmark &              \\
      REAL-FND~\cite{mosallanezhad2022domain} & \checkmark &           \\
      MTSD~\cite{li2021multi}     &              & \checkmark \\
       Llama-2-7B~\cite{touvron2023llama} & \checkmark & \checkmark          \\
      \midrule
      \textsc{\model} & \checkmark & \checkmark \\
      \bottomrule 
    \end{tabular}}
  \end{center}
    \vspace{-5pt}

\end{table}

\begin{table*}[!ht] \vspace{-15pt}
\centering
\footnotesize
\caption{Cross-domain, single-target performance results in accuracy and AUC (in parentheses). In this experiment we only use the datasets that have the same target word (p-value $<$ 0.05 for all McNemar’s tests).\label{tab:cross-domain-single-target}}
\resizebox{.8 \linewidth}{!}{\begin{tabular}{c|l|l|l|l|l|l|l|l}
\toprule
\multicolumn{1}{l|}{\textbf{Source Domain}} & \textbf{Target Domain} & \textbf{BERT} & \textbf{MoLE} & \textbf{MTL} & \textbf{MTSD} & \textbf{RLFND} & \textbf{Llama-2-7B} & \textbf{\model} \\
\midrule
\multirow{2}{*}{CoVaxNet\_pre} & CoVaxNet\_post & 0.713 (0.697) & 0.646 (0.621) & 0.734 (0.736) & 0.734 (0.731) & 0.752 (0.747) & 0.721 (0.725) & \textbf{0.765 (0.753)} \\
    & CoVaxNet\_pre & 0.762 (0.727) & \textbf{0.901 (0.876)} & 0.823 (0.789) & 0.845 (0.831) & 0.843 (0.835) & 0.812 (0.807) & 0.863 (0.858) \\ \midrule
\multicolumn{2}{c|}{\textbf{Performance Degradation}} & \textbf{0.049 (0.030)} & 0.255 (0.255) & 0.089 (0.053) & 0.111 (0.100) & 0.091 (0.088) & 0.091 (0.082)  & 0.098 (0.105) \\

\midrule \midrule
\multirow{2}{*}{CoVaxNet\_post} & CoVaxNet\_post & 0.826 (0.796) & 0.821 (0.835) & 0.811 (0.823) & \textbf{0.841 (0.839)} & 0.812 (0.784) & 0.761 (0.719)  & 0.831 (0.837) \\
    & CoVaxNet\_pre & 0.698 (0.655) & 0.713 (0.702) & 0.623 (0.611) & 0.712 (0.692) & 0.674 (0.632) & 0.688 (0.641)  & \textbf{0.723 (0.723)} \\ \midrule
\multicolumn{2}{c|}{\textbf{Performance Degradation}} & 0.128 (0.141) & 0.108 (0.133) & 0.188 (0.212) & 0.129 (0.147) & 0.138 (0.152) & \textbf{0.073 (0.078)}  & 0.108 (0.114) \\
\bottomrule
\end{tabular}}
\end{table*}

\begin{table*}[!ht]
\centering
\footnotesize
\caption{Cross-domain, cross-target performance results in accuracy and AUC (in parentheses). In this experiment, we evaluate the model's performance on a dataset that has a different target word in comparison to the source domain's target (p-value $<$ 0.05 for all McNemar’s tests).\label{tab:cross-domain-cross-target}}
\resizebox{.8 \linewidth}{!}{%
\begin{tabular}{c|l|l|l|l|l|l|l|l}
\toprule
\multicolumn{1}{l|}{\textbf{Source Domain}} & \textbf{Target Domain} & \textbf{BERT} & \textbf{MoLE} & \textbf{MTL} & \textbf{MTSD} & \textbf{RLFND} & \textbf{Llama-2-7B} & \textbf{\model} \\
\midrule
\multirow{4}{*}{CoVaxNet\_pre}  
    & Face Masks & 0.623 (0.568) & 0.732 (0.711) & 0.741 (0.712) & 0.789 (0.746) & 0.742 (0.625) & 0.862 (0.815) & \textbf{0.868 (0.823)} \\
    & Fauci & 0.741 (0.706) & 0.725 \textbf{(0.719)} & 0.691 (0.687) & 0.732 (0.698) & 0.699 (0.625) & 0.673 (0.635) & \textbf{0.743} (0.714) \\
    & School Closures & 0.832 (0.627) & 0.778 (0.761) & 0.711 (0.651) & 0.872 (0.731) & 0.834 (0.804) & 0.894 \textbf{(0.796)} & \textbf{0.920} (0.734) \\
    & Stay at Home & 0.641 (0.625) & 0.734 (0.738) & 0.621 (0.613) & 0.738 (0.732) & 0.665 (0.658) & \textbf{0.761 (0.759)} & 0.736 (0.742) \\ \midrule
\multicolumn{2}{c|}{\textbf{Average Performance}} & 0.709	(0.631) &	0.742	(0.732) &	0.691	(0.665) & 0.782	(0.726) & 0.735 (0.678) & 0.797 (0.751) & \textbf{0.816	(0.753)} \\
\midrule \midrule
\multirow{4}{*}{CoVaxNet\_post} 
    & Face Masks & 0.570 (0.555) & 0.801 \textbf{(0.786)} & 0.698 (0.676) & 0.794 (0.754) & 0.734 (0.712) & 0.808 (0.758) & \textbf{0.819} (0.769) \\
    & Fauci & 0.548 (0.624) & 0.721 (0.740) & 0.709 (0.683) &  \textbf{0.763 (0.778)} & 0.713 (0.694) & 0.678 (0.644) & 0.743 (0.714) \\
    & School Closures & 0.578 (0.543) & 0.718 (0.743) & 0.621 (0.620) & 0.687 (0.701) & 0.694 (0.675) & 0.725 (0.736) & \textbf{0.739 (0.772)} \\
    & Stay at Home & 0.623 (0.637) & 0.672 (0.691) & 0.669 (0.681) & 0.699 (0.704) & 0.675 (0.662) & 0.705 (0.711) & \textbf{0.725 (0.727)} \\\midrule
\multicolumn{2}{c|}{\textbf{Average Performance}} & 0.579 (0.589) & 0.728 (0.740) & 0.674 (0.665) &	0.735 (0.734) & 0.704 (0.685) & 0.729 (0.712) & \textbf{0.756 (0.745)} \\
\bottomrule
\end{tabular}}
\end{table*}

\subsection{Baselines}
We consider the following baselines. \autoref{tab:baselines} shows the difference between baselines and the proposed approach. Notably, we have implemented baselines for both cross-domain and cross-target tasks regardless of their specific task targets. This decision stems from the unique nature of the task, which presents limitations in terms of available baselines. We chose BERT as our base model instead of larger language models due to its demonstrated effectiveness in text classification, accessibility to task-specific pretrained models, and computational efficiency.

\noindent\textbf{MoLE~\cite{hardalov-etal-2021-cross}:} Utilizes mixture-of-experts models where each domain expert represents each domain and produces probabilities for all the target labels. Then uses domain adversarial training to learn domain-invariant representations. 

\noindent\textbf{MTL~\cite{schiller2021stance}:} A benchmarking framework for evaluating the robustness of stance detection systems. The proposed method leverages a diverse set of challenging datasets with varying levels of noise, bias, and adversarial attacks to evaluate the performance and robustness of stance detection systems. For each dataset, this model adds a domain-specific projection layer to the final layer of a pre-trained language model and freezes other layers during training. 

\noindent\textbf{RLFND~\cite{mosallanezhad2022domain}:} A domain-adaptive reinforcement learning framework for fake news detection. The proposed method leverages a deep reinforcement learning algorithm to learn the optimal policy for fake news detection while adapting to the domain shift between the source and target domains. \textbf{MTSD~\cite{li2021multi}:} A multi-task framework that performs stance detection on multiple targets. The framework leverages a shared encoder to capture the common features of the text and a task-specific decoder to predict the stance toward each target. 

\noindent\textbf{Llama-2-7B~\cite{touvron2023llama}:} A pre-trained and fine-tuned large language model released by Meta AI, designed for a variety of NLP tasks. We use the smallest variant for our problem setting. Due to Llama-2's incapability of performing stance detection on our dataset with an acceptable performance, NeMo framework was used to perform Supervised Fine-Tuning (SFT) on this model.

\subsection{Datasets}
We use two representative datasets related to COVID-19, each from different domains and targets (details provided in~\autoref{tab:table1}). The large-scale Twitter dataset CoVaxNet~\cite{jiang2022covaxnet} was employed for training across both domains and targets. CoVaxNet was compiled throughout the pandemic period (Jan 1, 2020 - Dec 31, 2021) and annotated with stances towards \emph{COVID-19 vaccination}. To evaluate the performance of \model~ in a single-target context, we categorized the \textit{domains} as distinct time intervals distinguished by significant events, following common practice in social media analysis~\cite{pak-paroubek-2010-twitter}. We divided CoVaxNet into two periods, with the full approval of the Pfizer-BioNTech vaccine by the US Food and Drug Administration (FDA) on Aug 23, 2021, serving as the dividing event. Thus, the source and target domains consist of 10,000 tweets each, selected randomly from January 1, 2020 to Aug 22, 2021 (``CoVaxNet\_pre''; source domain) and from Aug 23, 2021 to Dec 31, 2021 (``CoVaxNet\_post''; target domain). 

For the cross-target scenario, CoVaxNet\_pre and CoVaxNet\_post were separately used as training sets for \model. Model performance was evaluated on the targets (\emph{``Anthony S. Fauci, M.D.''}, \emph{``Wearing a Face Mask''}, \emph{``Keeping Schools Closed''}, \emph{``Stay at Home Orders''}) from the COVID-19-Stance~\cite{glandt2021stance} dataset\footnote{Dataset collected between Feb 27, 2020 and Aug 20, 2020}. It includes different targets and keywords of interest (e.g., \#lockdown), thereby justifying an adequate distributional shift between the source and target. Domain counterfactuals for this dataset were generated in the same manner as the cross-domain setting to augment the training set. Importantly, we labeled the generated examples the same as the original example as we observed a positive correlation between the targets in the source and target domain. Future works should explore scenarios where the stances towards targets in source and target domains are conflicting.

We excluded tweets that were too short or contained excessive hashtags, emojis, and URLs from both datasets to preserve their quality. Additionally, we balanced the training and test set across the different labels. It should be noted that congruent labels from the two datasets were treated as equivalent; that is, \textit{pro} is equivalent to \textit{in-favor} and \textit{anti} is the same as \textit{against}. In the context of stance detection, the terms in these pairs essentially convey the same stance. Both CoVaxNet and COVID-19-Stance datasets are publicly available for research purposes~\cite{jiang2022covaxnet}.


\normalsize

\subsection{Evaluation and Results}
\autoref{tab:cross-domain-single-target} and \autoref{tab:cross-domain-cross-target} summarize the comparative performance of the baseline and proposed models concerning the accuracy and AUC (in parentheses). 

\noindent\textbf{Q1:} 
In this experiment, the proposed approach and the domain-adaptive baselines used a fixed portion (30\%) of the target domain data during counterfactual data generation and maintained the same ratio in the training set. The models were trained on a single domain and tested on both source and target domains for stance detection for a single target of interest from the source domain. Our results provide some evidence that even when the target of interest is contained in the source domain, performance in the target domain generally degrades for all models. The \textit{Performance Degradation} in the table highlights how a model's performance changes when shifting from the source domain to the target domain. We observe that all models exhibit performance degradation, confirming the difference between domains.


\noindent\textbf{Q2:} 
The results of our experiments on the cross-domain cross-target setting, as presented in \autoref{tab:cross-domain-cross-target}, demonstrate the performance of our approach and the baselines. In this setting, we augment the training set of baselines with a small portion (30\%) of data from the target domain. The results of our experiments suggest that the proposed approach is more effective than the domain-adaptive baselines. Specifically, our approach surpasses all baselines regarding average accuracy and AUC score, underscoring its suitability for the cross-domain cross-target context. Furthermore, these results underscore the robustness of our approach to variations in the target domain, a strength attributable to our counterfactual data generation technique which is tailored to produce data that better represents the target domain.



\begin{figure}[!t]\vspace{-10pt}
    \centering
    \begin{subfigure}[b]{0.23\textwidth}
        \centering
        \includegraphics[width=\textwidth]{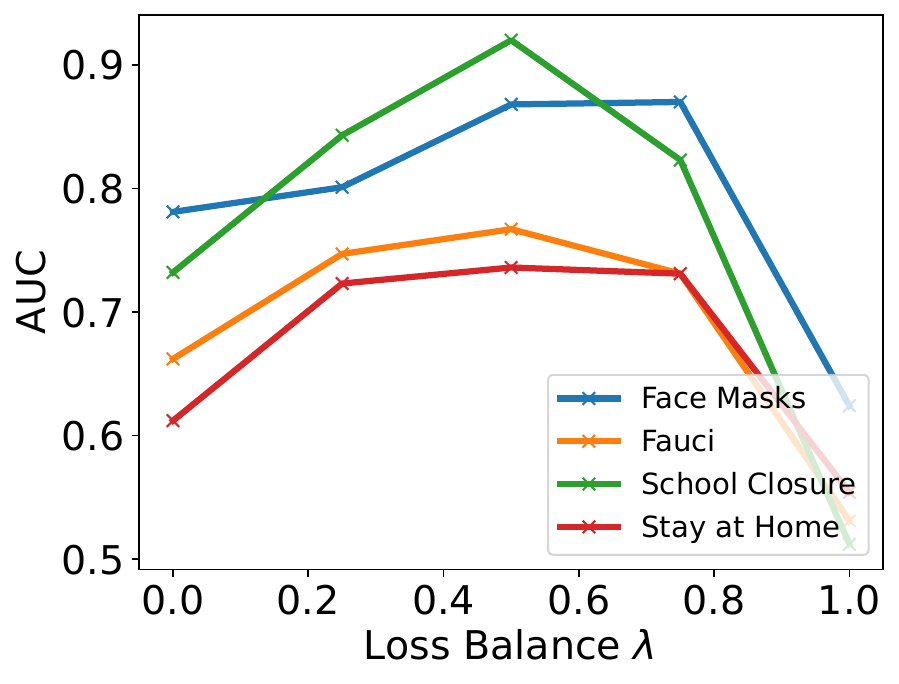}
        \caption{}
    \end{subfigure}
    \begin{subfigure}[b]{0.23\textwidth}
         \includegraphics[width=\textwidth]{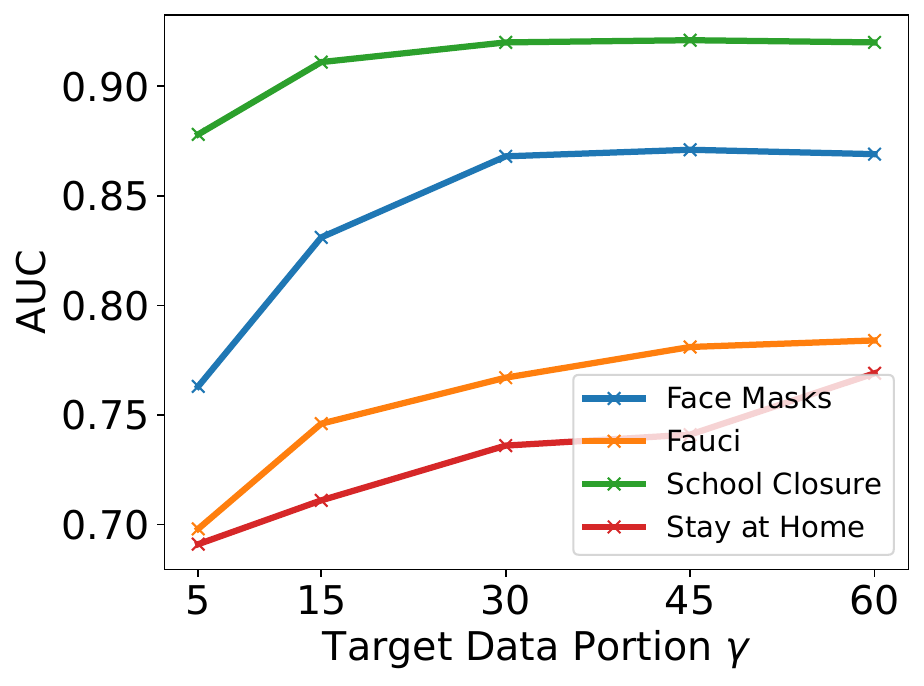}
         \caption{}
    \end{subfigure} \vspace{5pt}

    \begin{subfigure}[b]{0.25\textwidth}
         \includegraphics[width=\textwidth]{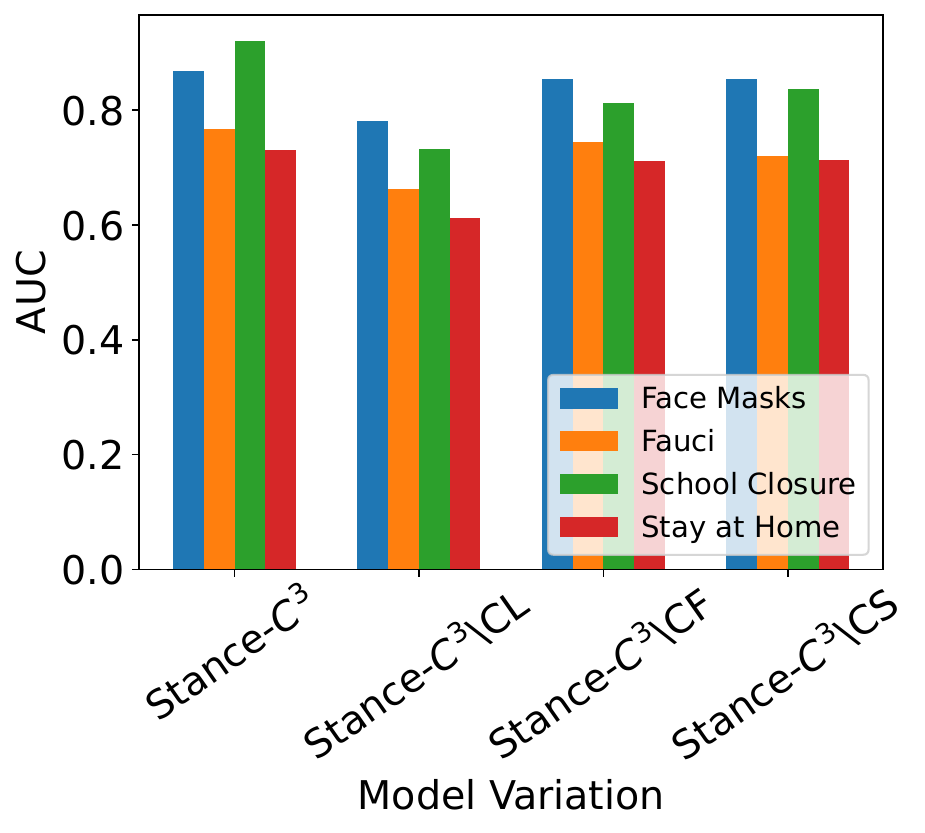}
         \caption{}
    \end{subfigure}
    \caption{{\footnotesize Impact analysis of model's parameters and components. Figures (a) and (b) show the impact of the target domain data portion and the balance between the loss values, respectively. Figure (c) shows the impact of different components - removing modified contrastive loss (\model \textbackslash CL), removing counterfactual data generation component (\model \textbackslash CF), and using simple contrastive loss (\model \textbackslash CS) - on the model's performance.\label{fig:parameter}}}
    \label{fig:components_20}\vspace{-10pt}
\end{figure}

\noindent\textbf{Parameter Analysis} To address Q3, we analyze the impact of hyperparameters on the performance of \model. To demonstrate the effect of contrastive loss and the balance between contrastive and cross-entropy loss, we vary the $\lambda$ value in the range of $\{0.0, 0.25, 0.5, 0.75, 1.0\}$ during training. As illustrated in \autoref{fig:parameter}-a, the AUC score varies significantly for different $\lambda$ values. The results suggest that using $\lambda=0.5$ yields the best performance.

We further investigate the impact of the target domain data portion $\gamma$ utilized for counterfactual data generation by varying its value within the range of $\{5\%,15\%,30\%,45\%\}$. The results, as illustrated in \autoref{fig:parameter}-b, demonstrate that incorporating a larger portion of target domain data leads to enhanced performance. However, in practical scenarios, access to substantial data in the target domain is often constrained. Considering there is no statistically significant performance difference observed between $\gamma=30\%$ and $\gamma=45\%$, we conclude that employing 30\% of the target domain data enables an acceptable level of performance.

\noindent\textbf{Ablation Study}
We investigate the impact of each component in \model by testing different model variants: (1) \textbf{\model\textbackslash CL} where we remove the contrastive loss component by setting $\lambda=1.0$, (2) \textbf{\model\textbackslash CF} where we remove the counterfactual data generation component and only use $30\%$ of target domain data ($\gamma=30\%$) in the training set, and (3) \textbf{\model\textbackslash CS} where we replace the modified contrastive loss with unmodified triplet loss. \autoref{fig:parameter}-c shows AUC degradation when domain counterfactual examples are removed, and when contrastive loss is either removed or modified during the training process. While removing or replacing either component (counterfactuals and contrastive loss) yielded worse AUC's, the contrastive loss component seems to have the greatest impact on \model's performance. Further comparisons with the unmodified triplet loss show gains with respect to the AUC, suggesting that using the constrastive loss function is a step in the right direction for enhancing the robustness of stance detection models.

\section{Conclusion}
\label{s:conclusion}
The effective use of social media data in facilitating effective public discourse and policy interventions requires agile and adaptable NLP models. Stance detection, as an NLP task, could be a potent tool in accurately gauging public and group-level support or oppositions for critical societal topics. The primary challenge addressed in this work is the difficulty in adapting a pre-trained model to different domains and targets of interest, which limits its utility in time-sensitive applications, such as using social media data to determine public opinions on non-pharmaceutical interventions during a pandemic. Our proposed approach uses counterfactual data generation for extended contexts in which data could be sparse. We showed through ablation studies that this approach improves stance detection across different domains and new targets. Further, we showed that using a modified contrastive loss function complements the data augmentation strategy with much less training data required for the target domain. The proposed architecture in this paper, which combines the two strategies, has shown evidence of outperforming existing SOTA methods in COVID datasets. For applications wherein specialized NLP models present distinct advantages over general purpose models, we showed that aggregating marginal gains by combining strategies is a step in the right direction for robust and adaptable stance detection.

\bibliographystyle{IEEEtran}
\bibliography{asonam24}


\end{document}